\documentclass[final,3p,times,twocolumn]{elsarticle}

\usepackage{amssymb}

\usepackage{graphicx}
\usepackage{bm}
\usepackage{amssymb,amsmath}
\usepackage{xcolor}
\usepackage{soul}

\biboptions{sort&compress}

\journal{Journal of Computational Science}

\begin{document}

\begin{frontmatter}

\title{Physics-Informed Echo State Networks}
%
%
\author[tummw,ias]{N.A.K. Doan}
\author[tummw]{W. Polifke}
\author[ias2,cam]{L. Magri}
\address[tummw]{Department of Mechanical Engineering, Technical University of Munich, Germany}
\address[ias]{Institute for Advanced Study, Technical University of Munich, Germany}
\address[ias2]{Institute for Advanced Study, Technical University of Munich, Germany (visiting)}
\address[cam]{Department of Engineering, University of Cambridge, United Kingdom}

\begin{abstract}

We propose a physics-informed Echo State Network (ESN) to predict the evolution of chaotic systems. Compared to conventional ESNs, the physics-informed ESNs are trained to solve supervised learning tasks while ensuring that their predictions do not violate physical laws. This is achieved by introducing an additional loss function during the training, which is based on the system's governing equations. The additional loss function penalizes non-physical predictions without the need of any additional training data. This approach is demonstrated on a chaotic Lorenz system and a truncation of the Charney-DeVore system. Compared to the conventional ESNs, the physics-informed ESNs improve the predictability horizon by about two Lyapunov times. This approach is also shown to be robust with regard to noise.
The proposed framework shows the potential of using machine learning combined with prior physical knowledge to improve the time-accurate prediction of chaotic dynamical systems.


\end{abstract}


\end{frontmatter}

\section{Introduction}

Over the past few years, there has been a rapid increase in the development of machine learning techniques, which have been applied with success to various disciplines, from image or speech recognition \citep{Krizhevsky2012,Hinton2012} to playing Go \citep{Silver2016}. However, the application of such methods to the study and forecasting of physical systems has only been recently explored, including some applications in the field of fluid dynamics \citep{Raissi2019a,Ling2016,Jaensch2017,Wu2018}.
One of the major challenges for using  machine learning algorithms for the study of complex physical systems is the prohibitive cost of data generation and acquisition for training \citep{Duraisamy2018,Raissi2019b}. However, in complex physical systems, there exists a large amount of prior knowledge, such as governing equations and conservation laws, which can be exploited to improve existing machine learning approaches. These hybrid approaches, called \emph{physics-informed machine learning} or \emph{theory-guided data science} \cite{Karpatne2017a}, have been applied with some success to flow-structure interaction problems \citep{Raissi2019a}, turbulence modelling \citep{Ling2016}, the solution of partial differential equations (PDEs) \citep{Raissi2019b}, cardiovascular flow modelling \citep{Kissas2020}, and  physics-based object tracking in computer vision \cite{Stewart2017}.

In this study, we propose an approach to combine physical knowledge with a machine learning algorithm to time-accurately forecast the evolution of chaotic dynamical systems. The machine learning tools we use are based on reservoir computing \citep{Lukosevicius2009}, in particular,  Echo State Networks (ESNs). ESNs are used here instead of more conventional recurrent neural networks (RNNs), like the Long-Short Term Memory unit, because ESNs proved particularly accurate in predicting chaotic dynamics for a longer time horizon than other machine learning networks \citep{Lukosevicius2009}. ESNs are also generally easier to train than other RNNs, and they have recently been used to predict the evolution of spatiotemporal chaotic systems \citep{Pathak2018a,Pathak2018}. In the present study, ESNs are augmented by physical constraints to accurately forecast the evolution of two prototypical chaotic systems, the Lorenz system \citep{Lorenz1963} and the Charney-DeVore system \citep{Crommelin2004}. The robustness of the proposed approach with regard to noise is also analysed. Compared to previous physics-informed machine learning approaches,  which mostly focused on identifying solutions of PDEs using feedforward neural networks \citep{Raissi2019a,Raissi2019b,Kissas2020}, the approach proposed here is applied on a form of RNN for the modeling of chaotic systems. The objective is to train the ESN in conjunction with physical knowledge to reproduce the dynamics of the original system for the ESN to be a digital twin of the real system.

Section \ref{sec:method} details the method used for the training and for forecasting the dynamical systems, both with conventional ESNs and the newly proposed physics-informed ESNs (PI-ESNs). Results are presented in section \ref{sec:results} and final comments are summarized in section \ref{sec:conclusion}.

\section{Methodology}
\label{sec:method}
The Echo State Network (ESN) approach presented in \citep{Lukosevicius2012} is used here. Given a training input signal $\bm{u}(n)$ of dimension $N_u$ and a desired known target output signal $\bm{y}(n)$ of dimension $N_y$, the ESN learns a model with output $\widehat{\bm{y}}(n)$ matching $\bm{y}(n)$. $n=1, ..., N_t$ is the number of time steps, and $N_t$ is the number of data points in the training dataset covering a time window from $0$ until $T=(N_t-1)\Delta t$. Here, where the forecasting of a dynamical system is under investigation, the desired output signal is equal to the input signal at the next time step, i.e., $\bm{y}(n) = \bm{u}(n+1) \in \mathbb{R}^{N_y}$.

The ESN is composed of a randomised high dimensional dynamical system, called a reservoir, whose states at time $n$ are represented by a vector, $\bm{x}(n) \in \mathbb{R}^{N_x}$ representing the reservoir neuron activations. The reservoir is coupled to the input signal, $\bm{u}$, via an input-to-reservoir matrix, $\bm{W}_{in} \in \mathbb{R}^{N_x\times N_u}$. The output of the reservoir, $\widehat{\bm{y}}$, is deduced from the states via the reservoir-to-output matrix, $\bm{W}_{out} \in \mathbb{R}^{N_y \times N_x}$, as a linear combination of the reservoir states:
\begin{equation}
    \widehat{\bm{y}}=\bm{W}_{out} \bm{x}
\end{equation}
In this work, a non-leaky reservoir is used, in which the state of the reservoir evolves according to:
\begin{equation}
\bm{x}(n+1) = \tanh \left( \bm{W}_{in} \bm{u}(n+1) + \bm{W} \bm{x}(n) \right)
\end{equation}
where $\bm{W} \in \mathbb{R}^{N_x\times N_x}$ is the recurrent weight matrix and the (element-wise) $\tanh$ function is used as an activation function for the reservoir neurons. The commonly-used $\tanh$ activation offers good accuracy \cite{Lukosevicius2009,Lukosevicius2012} for the systems studied here, as discussed in the results sections \ref{sec:Lorenz} and \ref{sec:CDV}. While different activation functions have been proposed \cite{Verzelli2019}, it is beyond the scope of the present work to study the effect of activation functions on the echo state network accuracy.

\begin{figure}[!ht]
	\centering
	\includegraphics[width=222pt]{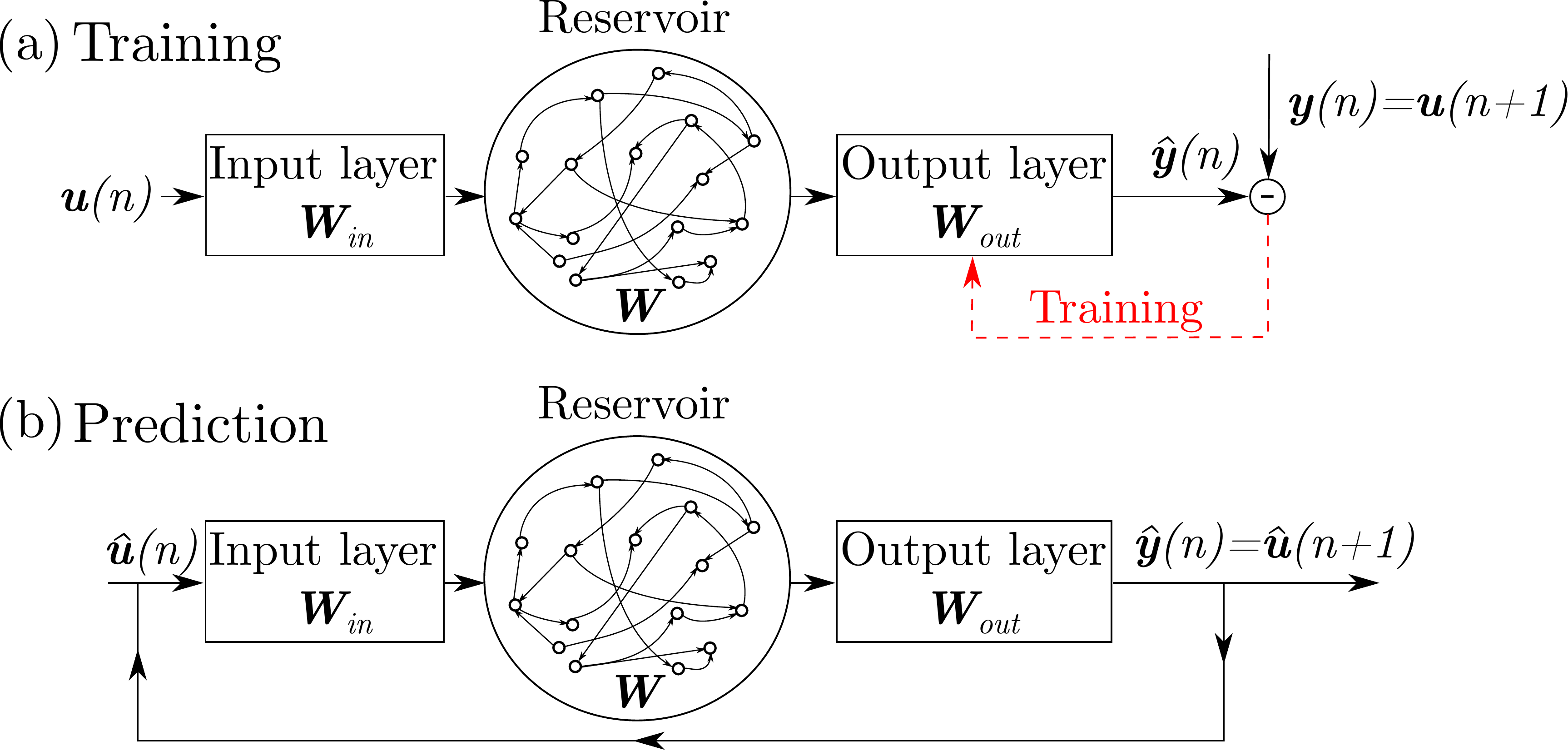}
	\caption{Schematic of the ESN during (a) training and (b) future prediction. The physical constraints are imposed during the training phase (a). }
	\label{fig:ESN_schema}
\end{figure}

In the conventional ESN approach (Fig. \ref{fig:ESN_schema}a), the input and recurrent matrices, $\bm{W}_{in}$ and $\bm{W}$, are randomly initialized only once and are not trained. These are typically sparse matrices constructed so that the reservoir verifies the Echo State Property \citep{Jaeger2004}. Only the output matrix, $\bm{W}_{out}$, is  trained to minimize the mean squared error, $E_d$, between the ESN predictions and the data:
\begin{equation}
E_{d} = \frac{1}{N_y} \sum_{i=1}^{N_y} \frac{1}{N_t} \sum_{n=1}^{N_t} (\widehat{y}_i (n) - y_i (n) )^2 \label{eq:err_data}
\end{equation}
(The subscript $d$ is used to indicate the error based on the available data.) Following \citep{Pathak2018a}, $\bm{W}_{in}$ is generated for each row of the matrix to have only one randomly chosen nonzero element, which is independently taken from a uniform distribution in the interval $[-\sigma_{in}, \sigma_{in}]$. $\bm{W}$ is constructed to have an average connectivity $\langle d \rangle$ and the non-zero elements are taken from a uniform distribution over the interval $[-1,1]$. All the coefficients of $\bm{W}$ are then multiplied by a constant coefficient for the largest absolute eigenvalue of $\bm{W}$ to be equal to a value $\Lambda$ where $\Lambda \leq 1$ to ensure the Echo State Property \citep{Lukosevicius2012}.

After training, to obtain predictions for future times $t>T$, the output of the ESN is looped back as an input, which evolves autonomously (Fig. \ref{fig:ESN_schema}b).

\subsection{Training}
Training of the ESN consists of the optimization of $\bm{W}_{out}$. As the outputs of the ESN, $\widehat{\bm{y}}$, are a linear combination of the states, $\bm{x}$, $\bm{W}_{out}$ can be obtained by using ridge regression:

\begin{equation}
\bm{W}_{out} = \bm{Y} \bm{X}^T \left( \bm{X} \bm{X}^T + \gamma \bm{I}  \right)^{-1} \label{eq:ridge}
\end{equation}
where $\bm{Y}$ and $\bm{X}$ are respectively the column-concatenation of the various time instants of the output data, $\bm{y}$, and associated ESN states $\bm{x}$. $\gamma$ is a Tikhonov regularization factor. The optimization in Eq. (\ref{eq:ridge}) is:

\begin{equation}
\bm{W}_{out} = \underset{\bm{W}_{out}}{\text{argmin}} \frac{1}{N_y} \sum_{i=1}^{N_y} \left( \sum_{n=1}^{N_t} (\widehat{y}_i(n) - y_i(n) )^2 + \gamma || \bm{w}_{out,i} ||^2 \right)
\end{equation}
where $\bm{w}_{out,i}$ denotes the $i$-th row of $\bm{W}_{out}$. This optimization problem penalizes large values of $\bm{W}_{out}$, which generally improves the feedback stability and  avoids overfitting \citep{Lukosevicius2012}.

In this work, following the approach of \citep{Raissi2019b} for artificial deep feedforward neural networks, we propose an alternative approach for training $\bm{W}_{out}$, which combines the data available with prior physical knowledge of the system under investigation. Let us first assume that the dynamical system is governed by the following nonlinear differential equation:
\begin{equation}
\mathcal{F}(\bm{y}) \equiv \dot{\bm{y}} + \mathcal{N} (\bm{y}) = 0
\label{eq:dynamic}
\end{equation}
where $\mathcal{F}$ is a general non-linear operator, $(\dot{~})$ is the time derivative and $\mathcal{N}$ is a nonlinear differential operator. Equation (\ref{eq:dynamic}) represents a formal equation describing the dynamics of a generic nonlinear system. The training phase can be reframed to make use of our knowledge of $\mathcal{F}$ by minimising the mean squared error, $E_d$, and a physical error, $E_p$, based on $\mathcal{F}$:
\begin{equation}
E_{tot} = E_d + E_p, \text{~where~} E_p = \frac{1}{N_y} \sum_{i=1}^{N_y}  \frac{1}{N_p} \sum_{p=1}^{N_p} | \mathcal{F}(\widehat{y_i}(n_p))|^2
\label{eq:Etot}
\end{equation}
Here, the set $\lbrace \widehat{\bm{y}} (n_p)  \rbrace_{p=1}^{N_p}$ denotes the ``collocation points" for $\mathcal{F}$, which are defined as a prediction horizon of $N_p$ datapoints obtained from the ESN covering the time period $(T+\Delta t) \leq t \leq (T+N_p \Delta t)$. Compared to the conventional approach where the regularization of $\bm{W}_{out}$ is based on avoiding extreme values of $\bm{W}_{out}$, the proposed method regularizes $\bm{W}_{out}$ by using the prior physical knowledge. Equation (\ref{eq:Etot}), which is a key equation, shows how to constrain the prior physical knowledge in the loss function.
Therefore, this procedure ensures that the ESN becomes predictive because of  data training and the ensuing prediction is consistent with the physics.
It is motivated by the fact that in many complex physical systems, the cost of data acquisition is prohibitive and thus, there are many instances where only a small amount of data is available for the training of neural networks. In this context, most existing machine learning approaches lack robustness. The proposed approach better leverages on the information content of the data that the recurrent neural network uses. 
The physics-informed framework is straightforward to implement because it only requires the evaluation of the residual, but it does not require the computation of the exact solution. Practically, the optimization of $\bm{W}_{out}$ is performed using the L-BFGS-B algorithm \cite{Byrd1995} with the $\bm{W}_{out}$ obtained by ridge regression (Eq. (\ref{eq:ridge})) as the initial guess.

\subsection{Hybrid ESN}
For a machine learning model comparison, the PI-ESN, which includes physical knowledge as a penalty term in the loss function, is compared to the hybrid approach of \cite{Pathak2018a}. The hybrid approach combines an ESN with an approximate model, which provides a one-step forward prediction that is fed both as an input into the ESN and directly into the output layer. The reservoir is excited by both the original input data and the prediction of the approximate model. The output layer, $\bm{W}_{out}$, is trained by blending the reservoir states and the prediction from the approximate model. This approach increases the size of the input and output layers of the ESN by the number of degrees of freedom in the approximate model. In \cite{Pathak2018a}, the approximate model was based on the same governing equations as the original system (with one of the coefficients being slightly altered), which doubles the size of the output and input layers. A similar approach will be carried out here for comparison.

\section{Results}
\label{sec:results}
\subsection{Lorenz system}
\label{sec:Lorenz}
The approach described in section \ref{sec:method} is applied for forecasting the chaotic evolution of the Lorenz system, which is governed by the following equations \citep{Lorenz1963}:
\begin{subequations}
\begin{align}
    \dot{u_1} &= \sigma (u_2-u_1)  \label{eq:Lorenz1} \\
    \dot{u_2} &= u_1 (\rho-u_3)-u_2  \label{eq:Lorenz2} \\
    \dot{u_3} &= u_1 u_2-\beta u_3 \label{eq:Lorenz3}
\end{align}
\label{eq:Lorenz}
\end{subequations}
where $\rho=28$, $\sigma = 10$ and $\beta=8/3$. These are the standard values of the Lorenz system that spawn a chaotic solution \citep{Lorenz1963}. The size of the training dataset is $N_t=1000$ and the timestep between two time instants is $\Delta t = 0.01$. This corresponds to roughly 10 Lyapunov times \cite{Strogatz1994}.

The parameters of the reservoir both for the conventional and PI-ESNs are: $\sigma_{in} = 0.15$, $\Lambda = 0.4$ and $\langle d \rangle = 3$. In the case of the conventional ESN, $\gamma = 0.0001$. These values of the hyperparameters are taken from previous studies \citep{Pathak2018,Pathak2018a}.

For the PI-ESN, a prediction horizon of $N_p=1000$ points is used and the physical error is estimated by discretizing Eq. (\ref{eq:Lorenz}) using an explicit Euler time-integration scheme. The choice of $N_p=1000$ is used to balance the error based on the data and the error based on the physical constraints. A balancing factor, similar to the Tikhonov regularisation factor, could potentially also be used to do this. However, the proposed method based on collocation points provide  additional information for the training of the PI-ESN as the physical residual has to be minimized at the collocation points. Increasing $N_p$ may be beneficial for the accuracy of the PI-ESN, but at the cost of a more computationally expensive training. Therefore, $N_p$ is chosen as a trade-off.

\begin{figure}[!ht]
	\centering
    \includegraphics[width=167pt]{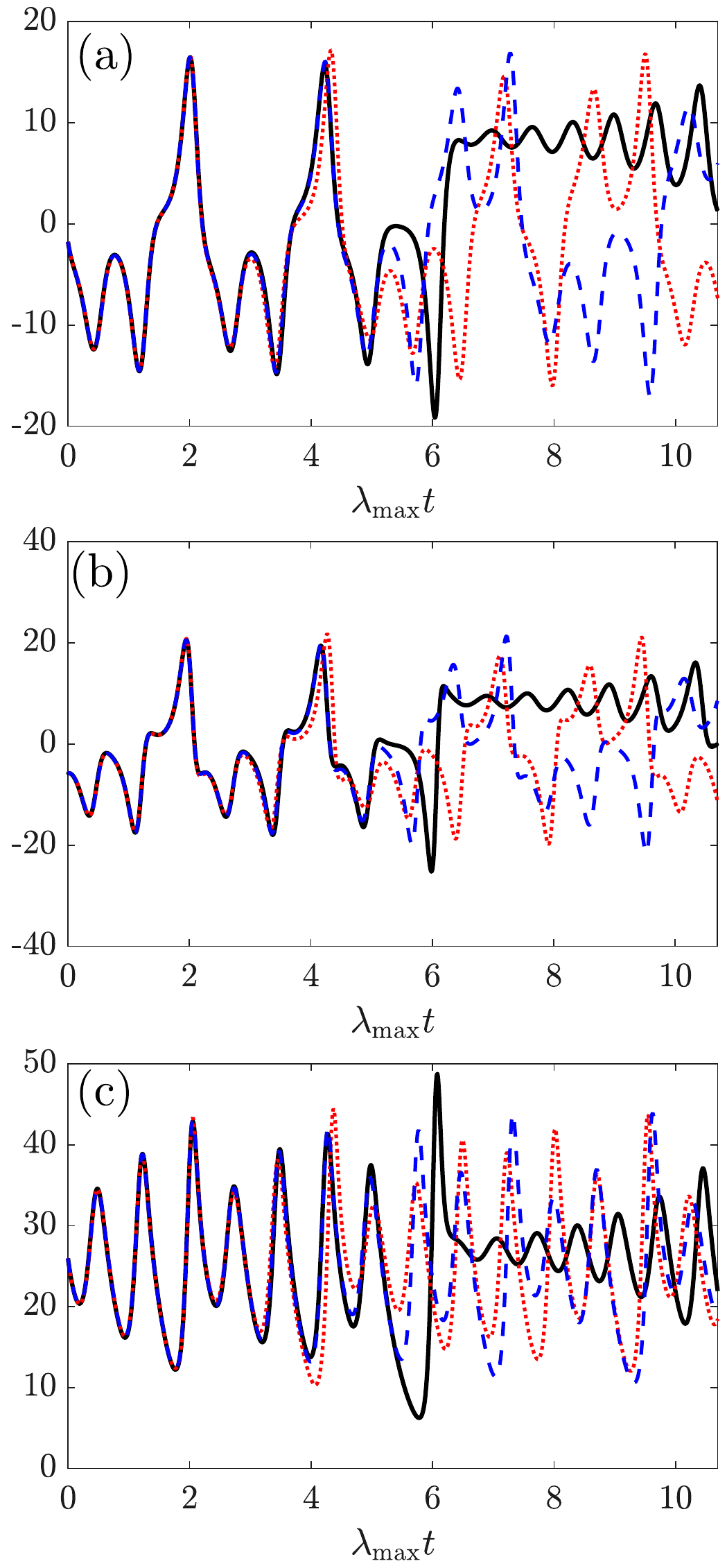}
	\caption{Prediction of the Lorenz system (a) $u_1$, (b) $u_2$, (c) $u_3$ with the conventional ESN (dotted red lines) and the PI-ESN (dashed blue lines). The actual evolution of the Lorenz system is shown with full black lines.}
	\label{fig:Lorenz_200U}
\end{figure}

\begin{figure}[!ht]
	\centering
	\includegraphics[width=177pt]{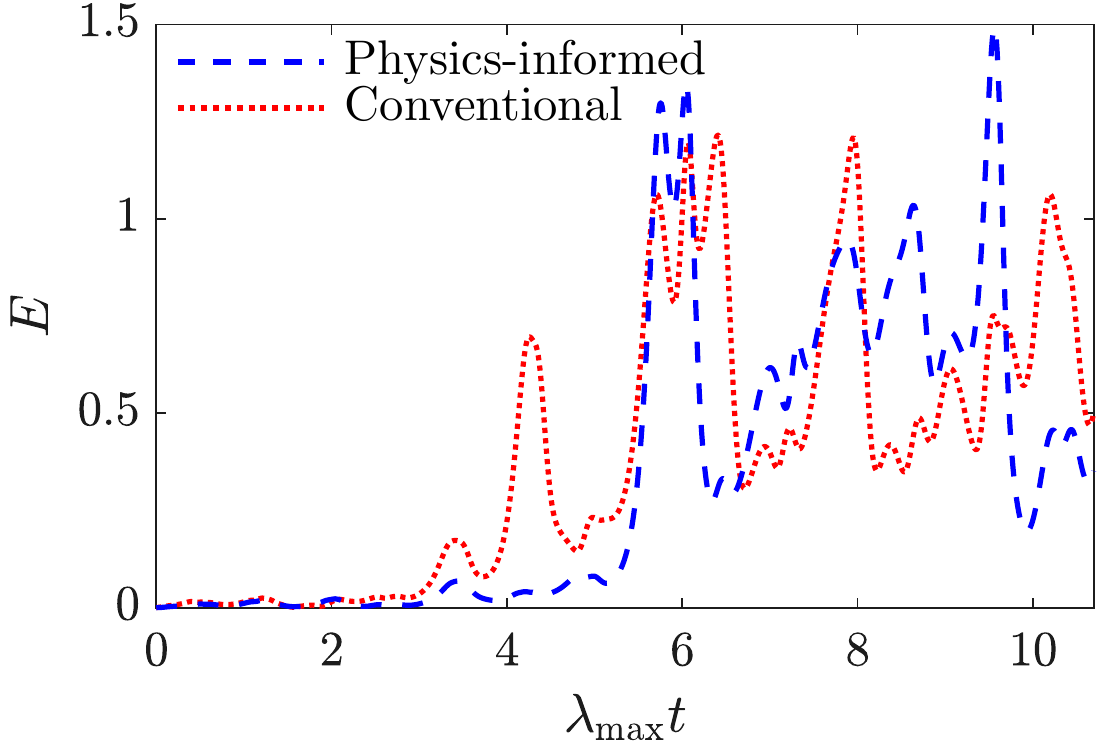}
	\caption{Error, $E$, from the conventional ESN (dotted red lines) and the PI-ESN (dashed blue lines) of the predictions shown in Fig. \ref{fig:Lorenz_200U}.}
	\label{fig:Lorenz_200U_E}
\end{figure}

The predictions for the Lorenz system by conventional ESN and PI-ESNs, for a particular case where the reservoir has 200 units, are compared with the actual evolution in Fig. \ref{fig:Lorenz_200U}, where the time is normalized by the largest Lyapunov exponent, $\lambda_{\max}=0.934$. Figure \ref{fig:Lorenz_200U_E} shows the evolution of the associated normalized error, which is defined as
\begin{equation}
E(n) = \frac{|| \bm{u}(n) - \widehat{\bm{u}}(n)||}{\langle || \bm{u} ||^2 \rangle^{1/2}}
\label{eq:error}
\end{equation}
where $\langle \cdot \rangle$ denotes the time average.
The PI-ESN shows a remarkable improvement of the time over which the predictions are accurate. Indeed, the time for the normalized error to exceed 0.2, which is the threshold used here to define the predictability horizon, increases from 4 Lyapunov times for the data-only ESN to 5.5 for the PI-ESN.

\begin{figure}[!ht]
	\centering
    \includegraphics[width=200pt]{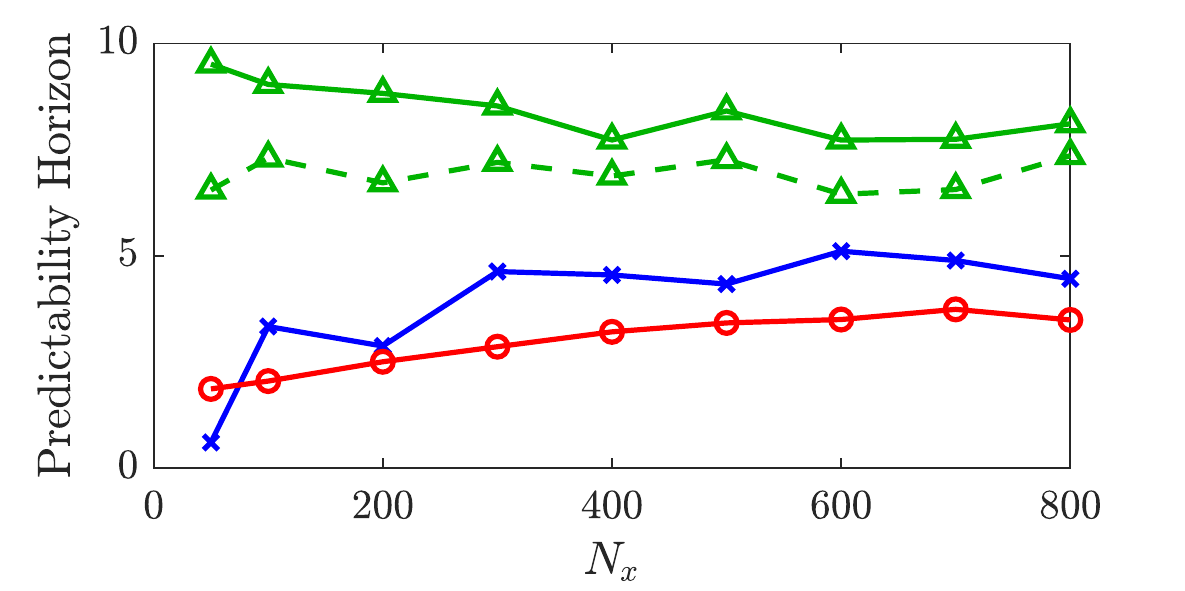} 
	\caption{Mean predictability horizon of the conventional ESN (red line with circles), PI-ESN (blue line with crosses), hybrid method with $\epsilon=0.05$ (green line with triangles) and hybrid method with $\epsilon=1.0$ (dashed green line with triangles) as a function of the reservoir size ($N_x$) for the Lorenz system.}
	\label{fig:valid_time}
\end{figure}

The statistical dependence of the predictability horizon on the reservoir size and the comparison with a hybrid ESN~\cite{Pathak2018a} are shown in Fig. \ref{fig:valid_time}. In the hybrid ESN, the approximate model consists of the same governing equations (Eqs. \eqref{eq:Lorenz}) with a slightly different parameter $\rho$, which is perturbed as $(1+\epsilon)\rho$ (as in~\cite{Pathak2018a}). Values of  $\epsilon=0.05$ and $\epsilon=1.0$ are considered here to have a higher- and lower-accuracy approximate model.
This statistical predictability horizon is estimated as follows. First, the trained PI-ESNs and conventional ESNs are run for an ensemble of 100 different initial conditions. Second, for each run, the predictability horizon is calculated. Third, the mean of the predictability horizon is computed from the ensemble.

It is observed that the physics-informed approach provides a marked improvement of the predictability horizon over conventional ESNs and, most significantly, for reservoirs of intermediate sizes. The only exception is for the smallest reservoir ($N_x=50$). 
In principle, it may be conjectured that a conventional ESN may have a similar performance to that of a PI-ESN by ad-hoc optimization of the hyperparameters. However, no efficient methods are available (to date) for hyperparameters optimization \citep{Lukosevicius2009}. The approach proposed here allows us to improve the performance of the ESN (optimizing $\bm{W}_{out}$) by adding a constraint on the physics, i.e., the governing equations, without changing the hyperparameters of the ESN and so, without performing an ad-hoc tuning of the hyperparameters. This suggests that the physics-informed approach may be more robust than the conventional approach and could provide an improvement of the accuracy of a given ESN without having to perform an expensive additional hyperparameter optimization.

The hybrid methods have a larger predictability horizon than both the PI-ESN and the conventional ESN with a downward, or constant, trend with increasing reservoir sizes. The hybrid model is more prone to overfitting as its output matrix is twice the size as the output matrix of the PI-ESN and conventional ESN. Furthermore, as it may be expected, the predictability horizon of the higher-accuracy case ($\epsilon=0.05$) is larger than the lower-accuracy case ($\epsilon=1.0$). The high predictability horizon of the hybrid ESN is due to the fact that the approximate solution is very close to the correct dynamics because the approximate model consists of the exact governing equations (with a small perturbation to a parameter).  Technically, the only difference between the approximate model and the governing equations is the $\epsilon{u_1}$ term in Eq. \eqref{eq:Lorenz2}. To compensate for this small error, the hybrid ESN does not need to learn the actual chaotic dynamics of the Lorenz system because the approximate model provides an accurate estimate. In practice, reduced-order models may contain larger model errors and may not model so accurately the actual dynamics of the system. Therefore, if the approximate solution is sufficiently far from the real dynamics, the gain in predictability horizon could become negligible despite the input and output layers being larger. This loss in accuracy in the hybrid ESN is illustrated in Sec. \ref{sec:results_noise} where noisy training data are considered. In contrast, the PI-ESN enables an improvement in predictability horizon without modifying the underlying network architecture.

\subsection{Charney-DeVore system}
\label{sec:CDV}
The truncated Charney-DeVore (CDV) system is now considered. This model is based on a Galerkin projection and truncation to 6 modes of the barotropic vorticity equation in a $\beta$-plane channel with orography \citep{Crommelin2004}. The 6 retained modes exhibit chaos and intermittency for an appropriate choice of parameters. The model equations are \citep{Wan2018,Crommelin2004,Crommelin2004a}:

\begin{align}
    \dot{u_1} &= \gamma_1^* u_3 - C(u_1 - u_1^*) \nonumber \\
    \dot{u_2} &= -(\alpha_1 u_1 - \beta_1)u_3 - Cu_2 - \delta_1 u_4 u_6 \nonumber \\
    \dot{u_3} &= (\alpha_1u_1 - \beta_1)u_2 - \gamma_1 u_1 -C u_3 + \delta_1 u_4 u_5 \nonumber \\
    \dot{u_4} &= \gamma_2^* u_6 - C(u_4 - u_4^*) + \epsilon (u_2 u_6 - u_3 u_5) \nonumber \\
    \dot{u_5} &= -(\alpha_2 u_1 - \beta_2) u_6 - C u_5 - \delta_2 u_4 u_3 \nonumber \\
    \dot{u_6} &= (\alpha_2 u_1 - \beta_2) u_5 -\gamma_2 u_4 - C u_6 + \delta_2 u_4 u_2
    \label{eq:CDV_eq}
\end{align}

where the model coefficients are given by:
\begin{align}
    \alpha_m = \frac{8 \sqrt{2}m^2(b^2 + m^2 -1)}{\pi(4m^2-1)(b^2+m^2)}, \hspace{11pt} \beta_m = \frac{\beta b^2}{b^2 + m^2} \nonumber \\
    \delta_m = \frac{64\sqrt{2}}{15\pi}  \frac{b^2-m^2+1}{b^2+m^2}, \hspace{11pt} \gamma_m^* = \gamma \frac{4\sqrt{2}mb}{\pi(4m^2-1)} \nonumber \\
    \epsilon = \frac{16\sqrt{2}}{5\pi}, \hspace{11pt} \gamma_m = \gamma \frac{4\sqrt{2}m^3b}{\pi(4m^2-1)(b^2+m^2)} \label{eq:CDV_coef}
\end{align}
for $m=1,2$. Here, we set the parameters as in \citep{Wan2018}, $(u_1^*,u_4^*,C,\beta,\gamma,b)=(0.95,-0.76095,0.1,1.25,0.2,0.5)$, which ensures a chaotic and intermittent behaviour.

\begin{figure}[!ht]
    \centering
    \includegraphics[width=177pt]{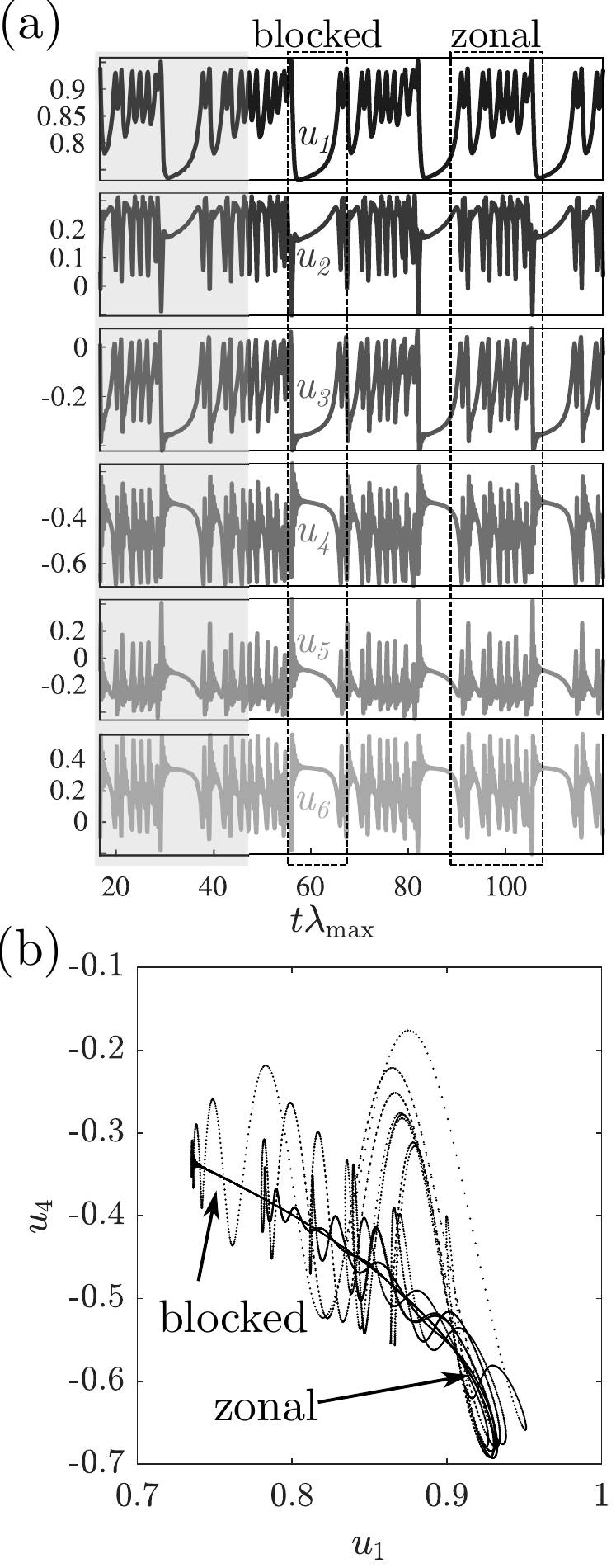}
    \caption{(a) Evolution of the modal amplitudes of the CDV system (black to light gray: $u_1$ to $u_6$). The shaded grey box indicates the data used for training. (b) Phase plots of the $u_1-u_4$ trajectory.}
    \label{fig:CDV_evol}
\end{figure}

The time evolution of this system is illustrated in Fig. \ref{fig:CDV_evol}. It can be seen that the CDV system shows two distinct regimes: one characterised by a slow evolution (and a large decrease in $u_1$) and one with strong fluctuations of all modes. These correspond to ``blocked" and ``zonal" flow regimes, respectively, which originate from the combination of topographic and barotropic instabilities \citep{Crommelin2004}. This intermittent characteristic of the CDV system makes it significantly more challenging than the Lorenz system. The dataset illustrated in Fig. \ref{fig:CDV_evol} is obtained by discretizing the set of equations (\ref{eq:CDV_eq}) with an Euler-explicit scheme with a timestep of $\Delta t = 0.1$. The first 9000 timesteps of Fig. \ref{fig:CDV_evol}, highlighted in the grey box, are kept for training. This corresponds to approximately 30 Lyapunov times. The largest Lyapunov exponent of the CDV system is equal to $\lambda_{\max}=0.033791$.

For the prediction, the parameters for the ESNs are: $\sigma_{in} = 2.0$, $\Lambda = 0.9$ and $\langle d \rangle = 3$. For the conventional ESN, $\gamma = 0.0001$. These values are obtained after performing a grid search. For the PI-ESN, a prediction horizon of $N_p=3000$ points is used. Compared to the Lorenz system where the same number of collocation points as training points was used, here, comparatively fewer collocation points are used. This choice was made to decrease the computational cost of the optimization process as the cost of computing $E_p$  is proportional to $N_p$. Nonetheless, that number of collocation points was sufficient to improve the prediction as is shown next.

\begin{figure}[!ht]
    \centering
    \includegraphics[width=200pt]{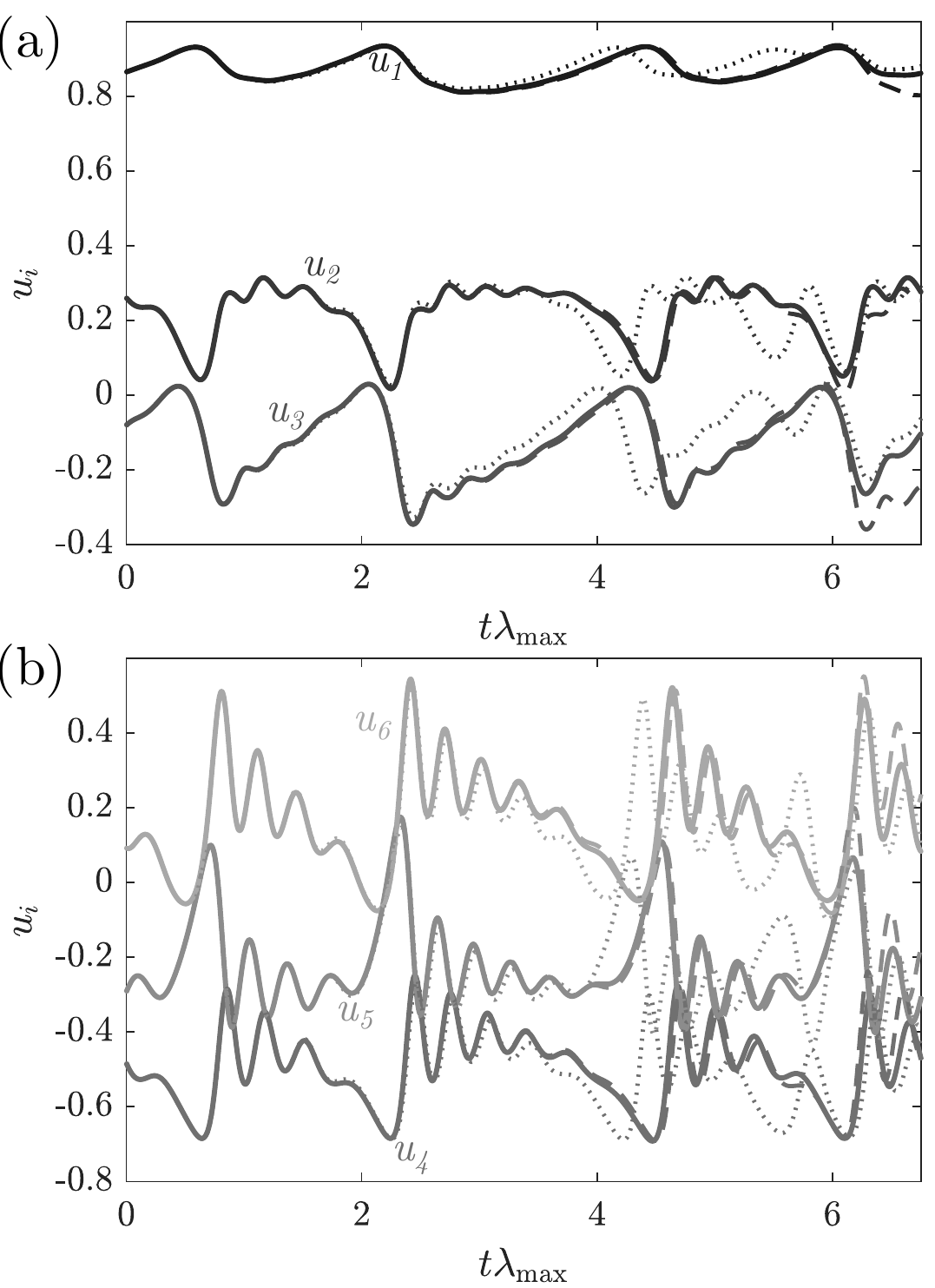}
    \caption{Prediction of the CDV system for (a) $u_1,~u_2$ and $u_3$ and (b) $u_4,~u_5$ and $u_6$ with the conventional ESN (dotted lines) and the PI-ESN (dashed lines). The actual evolution of the CDV system is shown with full lines.}
    \label{fig:CDV_pred}
\end{figure}

\begin{figure}[!ht]
    \centering
    \includegraphics[width=200pt]{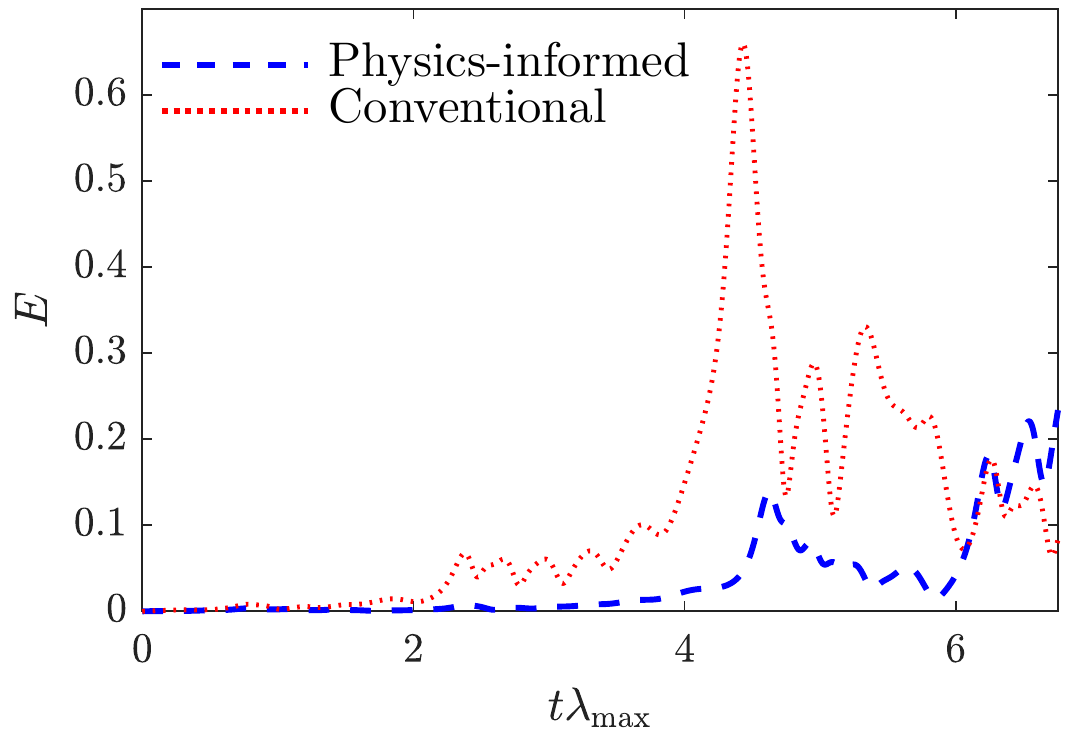}
    \caption{Error on the prediction from the conventional and PI-ESN for the prediction shown in Fig. \ref{fig:CDV_pred}.}
    \label{fig:CDV_pred_E}
\end{figure}

In Fig. \ref{fig:CDV_pred}, the predictions of the evolution of the CDV system by the ESN and PI-ESN with a reservoir of 600 units are presented alongside the true evolution. The associated normalised error (Eq. (\ref{eq:error})), is shown in Fig. \ref{fig:CDV_pred_E}. The PI-ESN outperforms the conventional ESN and maintains a good accuracy for 2 Lyapunov times beyond the conventional ESN. 

\begin{figure}[!ht]
    \centering
    \includegraphics[width=200pt]{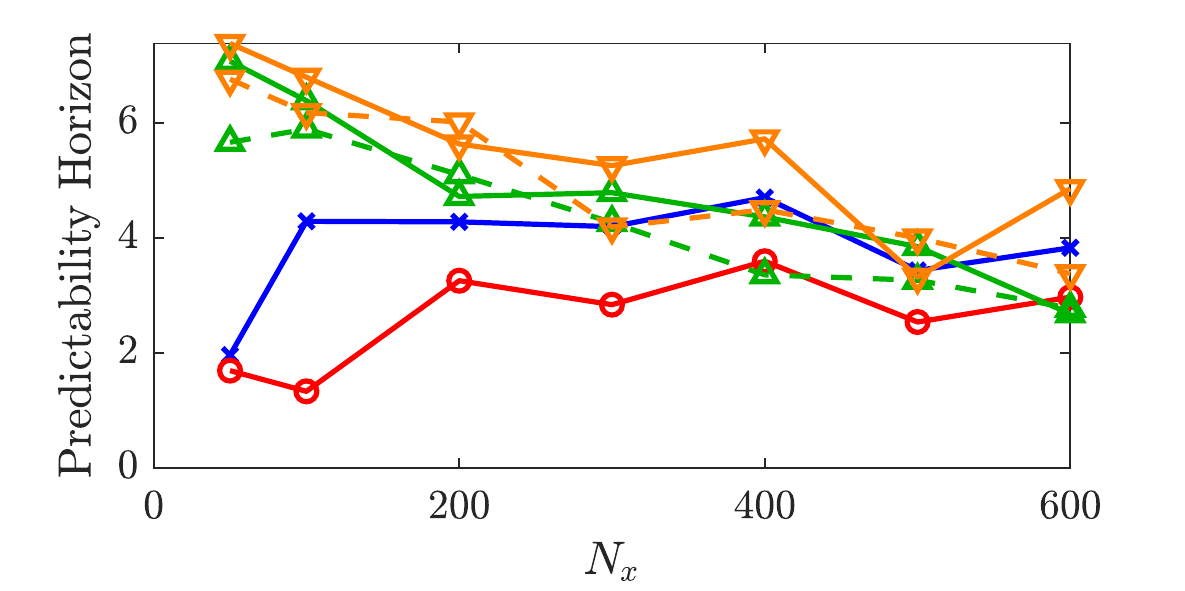}
    \caption{Mean predictability horizon of the conventional ESN (red line with circles), PI-ESN (blue line with crosses), hybrid-$b$ with $\epsilon=0.05$ (full green line with triangles), hybrid-$b$ with $\epsilon=1.0$ (dashed green line with triangles), hybrid-$C$ with $\epsilon=0.05$ (full orange line with downward triangles) and hybrid-$C$ with $\epsilon=1.0$ (dashed orange line with downward triangles) as a function of the reservoir size ($N_x$) for the CDV system.}
    \label{fig:CDV_valid_time}
\end{figure}

To assess the robustness of the results and compare the PI-ESN with the hybrid ESN, a statistical analysis similar to Sec. \ref{sec:Lorenz} is shown in Fig. \ref{fig:CDV_valid_time}. Similarly to the Lorenz system, the approximate model used consists of the exact governing equations (Eqs. \eqref{eq:CDV_eq}) with one  parameter being slightly perturbed. Two cases are considered: one in which $b$ is perturbed as  $(1+\epsilon)b$ (hybrid-$b$), and one in which $C$ is perturbed as $(1+\epsilon)C$ (hybrid-$C$), where $\epsilon=0.05$ or $1.0$.

The mean predictability horizon is computed from 100 different initial conditions and for different reservoir sizes. Similarly to the Lorenz system, the PI-ESN outperforms the conventional ESN by up to 2 Lyapunov times. However, the evolution of the predictability horizon of the PI-ESN and also the conventional ESN shows some degradation for very large reservoirs.
It is conjectured that this behaviour originates from overfitting and the more complicated evolution of the CDV system which exhibits two different regimes. Indeed, for the PI-ESN, the training is performed using the training timeseries which contains mostly a zonal regime evolution and the collocation points which are at times corresponding to a zonal regime as they are directly after the training dataset. As a result, the conventional ESN and the PI-ESN with very large reservoir may be overfitting to predict only the zonal regime. It is possible that by extending the collocation points for the PI-ESN, the prediction of the PI-ESN improves as those added collocation points may then cover a blocked regime evolution.

The hybrid ESN has a larger predictability horizon for small reservoirs because of the  extra information added by the approximate model, which is close to the exact model. The accuracy is, however, less marked than it is in the Lorenz system because an error in the parameters $b$ or $C$ is amplified by more significant model nonlinearities as these parameters appear in all the governing equations of the CDV system (Eq. \eqref{eq:CDV_eq}). The accuracy of hybrid-$b$  is lower than  the accuracy of hybrid-$C$  because the nonlinear dynamics is more sensitive to small errors in $b$, which affects all the coefficients of the CDV equations (Eqs. \eqref{eq:CDV_eq}-\eqref{eq:CDV_coef}). Similarly to the Lorenz system, when the model error is larger ($\epsilon=1.0$), the predictability horizon is smaller than with the accurate approximate model ($\epsilon=0.05$).

\subsection{Robustness with respect to noise}
\label{sec:results_noise}
In this section, we study the robustness of the results presented in the previous sections for the Lorenz and CDV systems with regard to noise.
To do so, the training data used in Sects. \ref{sec:Lorenz} and \ref{sec:CDV} are perturbed by adding measurement Gaussian noise to the training datasets. Two cases with Signal to Noise Ratios (SNRs) of 20 and 30dB are considered, which are typical noise levels encountered in experimental fluid mechanics \cite{Ouellette2006}.

\begin{figure}[!ht]
    \centering
    \includegraphics[width=177pt]{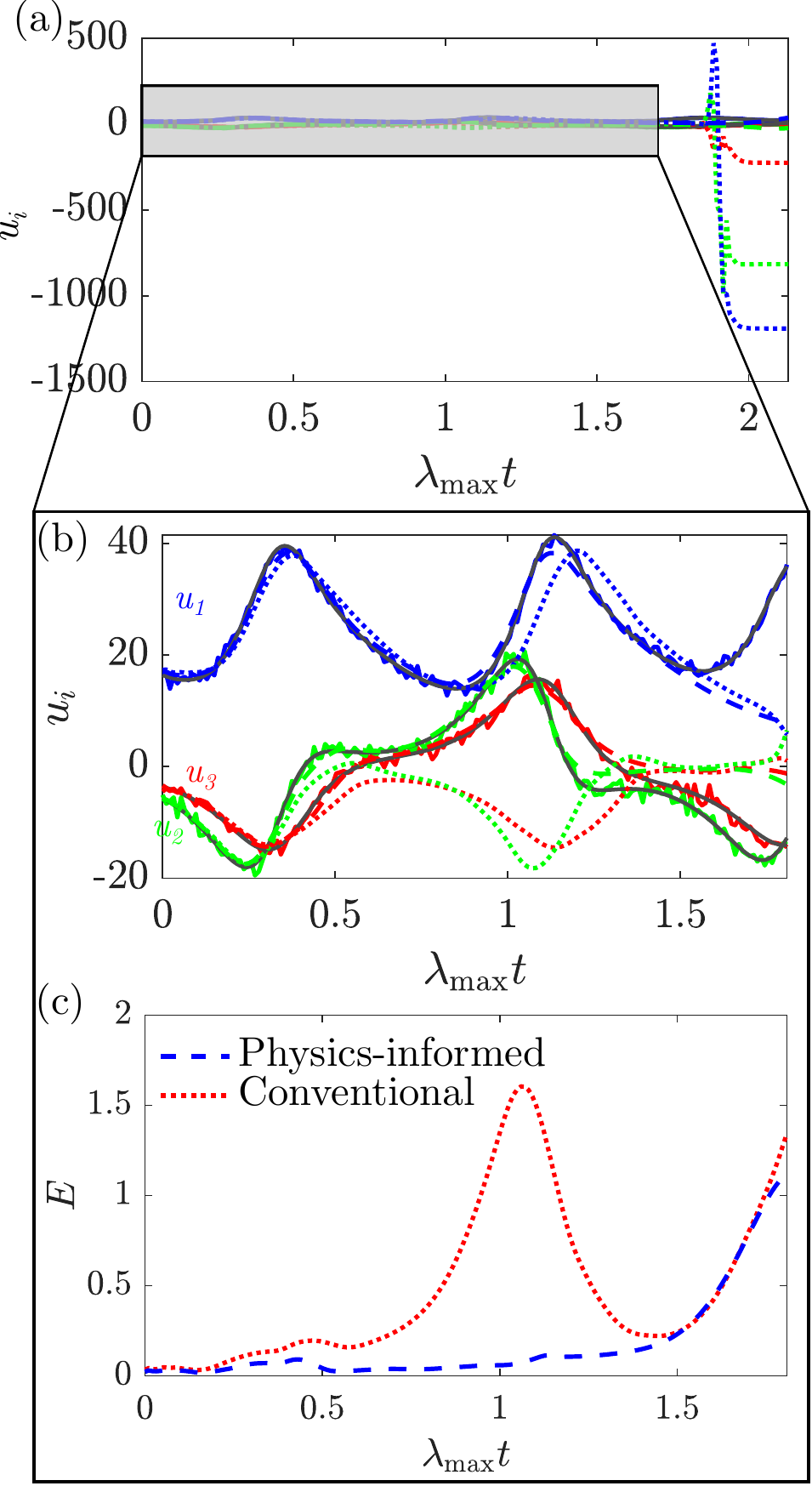}
    \caption{(a) Prediction of the Lorenz system with the conventional ESN (dotted lines) and the PI-ESN (dashed lines) with 200 units trained from noisy data (SNR=20dB) and (b) Zoom of the evolution before the divergence of the conventional ESN. The actual (noise-free) evolution of the Lorenz system is shown with full grey lines. (c) Error on the prediction for the conventional ESN and PI-ESN.}
    \label{fig:Lorenz_evol_noise}
\end{figure}

\begin{figure}[!ht]
    \centering
    \includegraphics[width=177pt]{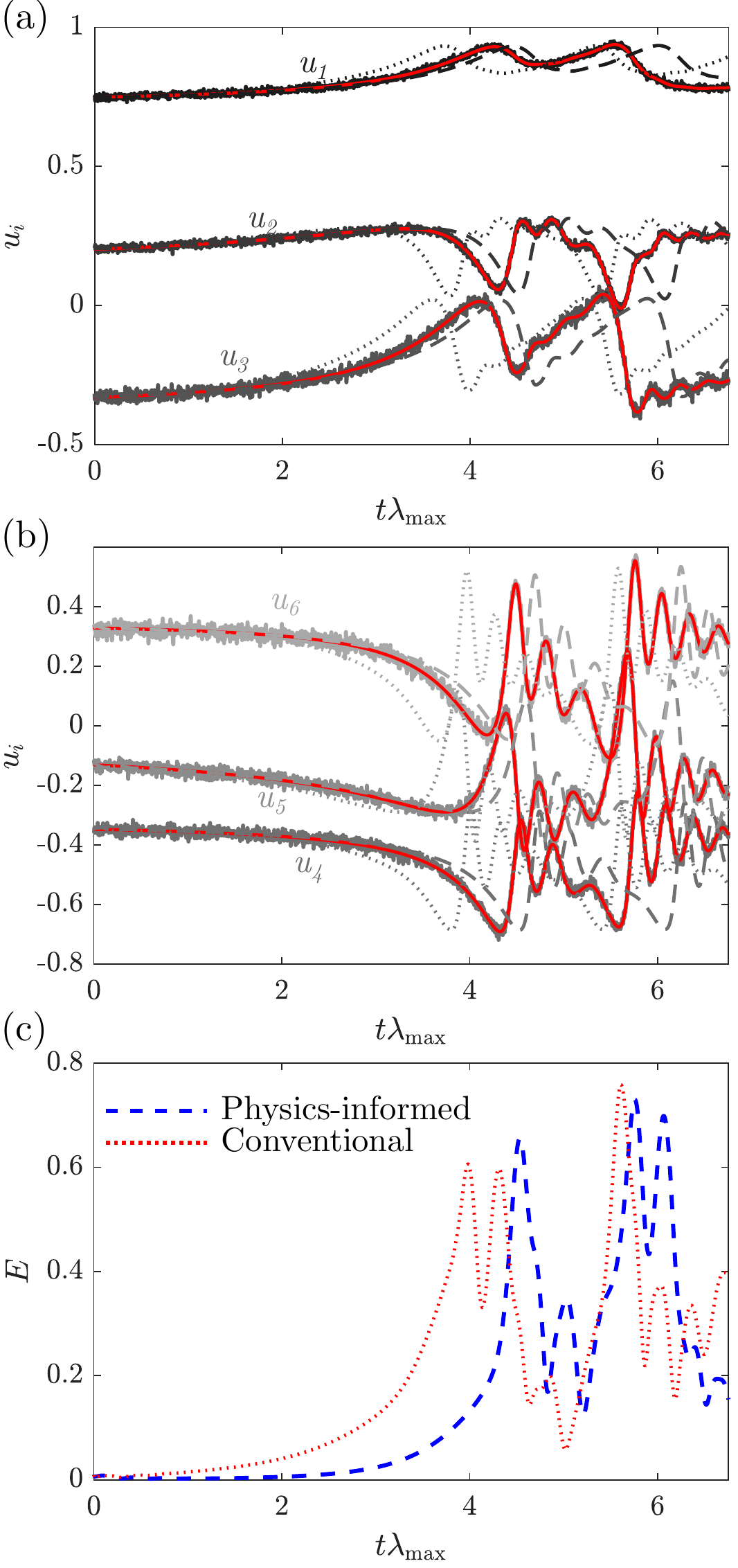}
    \caption{(a-b) Prediction of the CDV system with the conventional ESN (dotted lines) and the PI-ESN (dashed lines) with 600 units trained from noisy data (SNR=20dB). The actual (noise-free) evolution of the CDV system is shown with full red lines and the noisy data is shown with full greyscale lines. (c) Error on the prediction from the conventional ESN and PI-ESN.}
    \label{fig:CDV_evol_noise}
\end{figure}

The evolution of the Lorenz and the CDV systems and the predictions from the conventional and PI-ESNs are shown in Figs. \ref{fig:Lorenz_evol_noise} and \ref{fig:CDV_evol_noise}, respectively. In those figures, it is seen that the proposed approach still improves the prediction capability of the PI-ESN despite the training with noisy data.
This originates from the physics-based regularization term in the loss function in Eq. (\ref{eq:Etot}), which provides the information required during the training as to how to appropriately filter the noise. Indeed, the physics-based loss provides the constraints that the components of the output have to satisfy, therefore providing an indication as to how to filter the noise. In addition, for the Lorenz system, the conventional ESN is diverging during its prediction while the PI-ESN's prediction remains bounded. This highlights the improved robustness of the physics-informed approach.
This is an encouraging result, which can potentially enable the use of the proposed approach with noisy data from physical experiments whose governing equations are known.

\begin{figure}[!ht]
    \centering
    \includegraphics[width=200pt]{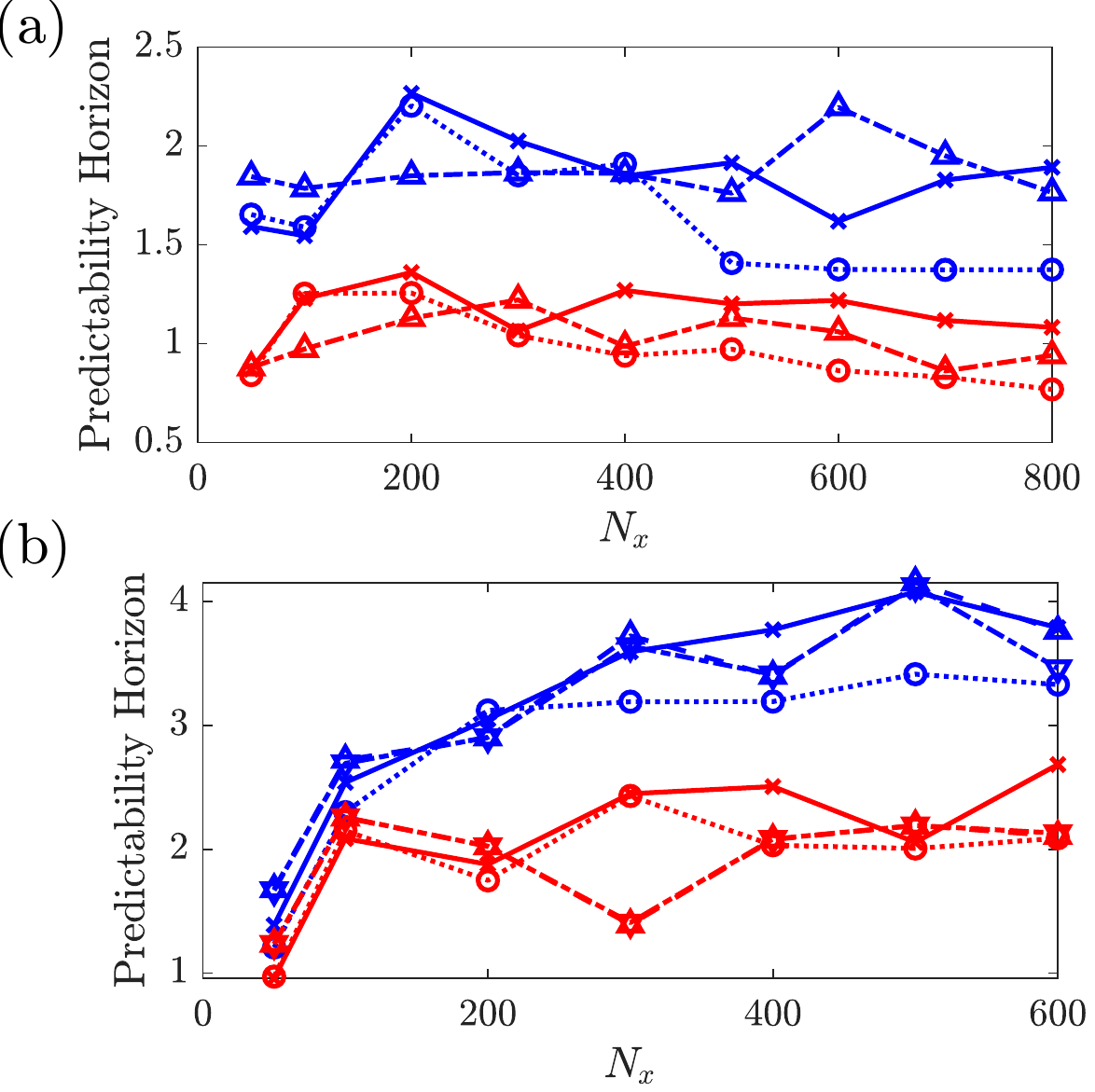}
    \caption{Mean predictability horizon of the conventional ESN (dotted line with circles), PI-ESN (full line with crosses), hybrid or hybrid-$b$ (dashed-dotted line with upward triangles) and hybrid-$C$ (dashed line with downward triangles) trained from noisy data (red: SNR=20dB, blue: SNR=30dB) as a function of the reservoir size ($N_x$) for the (a) Lorenz and (b) CDV systems. Hybrid methods are used with $\epsilon=0.05$}
    \label{fig:Lorenz_CDV_hor_time_noise}
\end{figure}

The mean predictability horizon for the two systems and the two noise levels is shown in Fig. \ref{fig:Lorenz_CDV_hor_time_noise}, which also shows a comparison with the hybrid approach with $\epsilon=0.05$. For the Lorenz system, compared to the ESN trained on non-noisy data, in Fig. \ref{fig:valid_time}, the mean predictability horizon is smaller. Furthermore, for the data-only ESN, the predictability horizon  decreases for large reservoirs. This is because the ESN starts overfitting the noisy data and, thereby, reproducing a noisy behaviour and deteriorating its prediction. On the other hand, the PI-ESN maintains a satisfactory predictability horizon for the same large reservoirs. This indicates that the physics-based regularization in the loss function ($E_p$ in Eq. (\ref{eq:Etot})) enhances the robustness of the PI-ESN.
The predictability horizon of the hybrid method  is  close to the predictability horizon of the PI-ESN for a small noise level.
This is due to the effect of noise in the training data. During the training, the approximate model time-integrates noisy input data, therefore, the approximate prediction is far from the target output. As a result, during the training, the hybrid ESN learns to rely mostly on the reservoir states to make a forecast, and only to use the prediction from the approximate model in a limited way. This is more apparent for a higher noise level, in which the predictability horizon of the hybrid method becomes shorter than the predictability horizon of the PI-ESN. This shows that the hybrid ESN is not filtering out the noise as efficiently as the PI-ESN.
The performance of the hybrid ESN deteriorates for a higher noise level.

For the CDV system, similar observations as for the Lorenz system can be made. However, the decrease in mean predictability horizon of the ESN and PI-ESN with large reservoir sizes is not observed as it has a larger dimension than the Lorenz system. Hence, it would require larger reservoirs than those considered here before the occurrence of noise overfitting.
Finally, the accuracy of the hybrid method is similar to that of the PI-ESN. Similarly to the Lorenz system, this is because of the effect of noisy data used in training.

\section{Conclusions and future directions}
\label{sec:conclusion}
In this paper, we propose an approach for training echo state networks (ESNs) by constraining the knowledge of the physical equations that govern a dynamical system. This physics-informed ESN (PI-ESN) is shown to be more robust than purely data-trained ESNs. The proposed PI-ESN needs minimal modification of the original architecture by requiring only the estimation of the physical residual. The predictability horizon is markedly increased without requiring additional training data. This is assessed on the Lorenz system and the Charney-DeVore system, both of which exhibit strong intermittency.
Furthermore, the robustness to noise of the proposed PI-ESN is assessed. It is observed that, compared to a Thikonov regularization, the PI-ESN performs more robustly, even with larger reservoirs where the conventional ESN may overfit the noisy data. As compared to other nonlinear filters used for denoising, such as the ensemble Kalman filter, the proposed approach does not require ensemble calculations.

For noise-free data, the predictability of the hybrid ESN~\cite{Pathak2018a} can be higher than the predictability of the PI-ESN, but the model errors of the additional approximate model in the hybrid ESN, which requires an additional time-integration, should be very small. In engineering practice, we expect model errors to be more significant. Additionally, the hybrid method needs larger output and input layers, up to twice the original size if the approximate model has the same number of states as the original system as in \cite{Pathak2018a}, and a time integrator for the approximate model.
For noisy data, the predictability of the PI-ESN is higher than the predictability of the hybrid method of~\cite{Pathak2018a}.

In addition, in ongoing work, the PI-ESN is being applied to high dimensional fluid dynamics systems. 
This work opens up new possibilities for the time-accurate prediction of the dynamics of chaotic systems by using the underlying physical laws as constraints.

\section*{Acknowledgements}
The authors acknowledge the support of the Technical University of Munich - Institute for Advanced Study, funded by the German Excellence Initiative and the European Union Seventh Framework Programme under grant agreement no. 291763. L.M. also  acknowledges the Royal Academy of Engineering Research Fellowship Scheme.

%


\end{document}